\documentstyle[epsf,named,ml98]{article} 

\begin{document}

\title{Top-down induction of clustering trees}
\author{Hendrik Blockeel \And Luc De Raedt\\
Katholieke Universiteit Leuven, Department of Computer Science\\
Celestijnenlaan 200A, B-3001 Heverlee, Belgium \\
\{Hendrik.Blockeel,Luc.DeRaedt,Jan.Ramon\}@cs.kuleuven.ac.be\\
\And Jan Ramon\thanks{~~The authors
are listed in alphabetical order.}}

\newcommand{\Tilde}{{\sc Tilde}}
\newcommand{\tich}{{\sc TIC}}
\newcommand{\Tic}{{\sc TIC}}

\maketitle

\begin{abstract}
  An approach to clustering is presented that adapts the basic
  top-down induction of decision trees method towards clustering.  To
  this aim, it employs the principles of instance based learning.  The
  resulting methodology is implemented in the TIC (Top down Induction
  of Clustering trees) system for first order clustering. The TIC
  system employs the first order logical decision tree representation
  of the inductive logic programming system \Tilde.  Various
  experiments with TIC are presented, in both propositional and
  relational domains.
\end{abstract}


\section{INTRODUCTION}
Decision trees are usually regarded as representing theories
for classification. The leaves of the tree contain the classes
and the branches from the root to a leaf contain sufficient conditions 
for classification.

A different viewpoint is taken in {\em Elements of Machine
  Learning} \cite{Langley96:other}.  
According to Langley, each  node of a tree
corresponds to a concept or a cluster,
and the tree as a whole thus represents 
a kind of taxonomy or a hierarchy.
Such taxonomies are not only output 
by decision tree algorithms
but typically also by clustering algorithms
such as e.g. COBWEB \cite{Fisher87:jrnl}.  
Therefore, Langley views both clustering and concept-learning as 
instantiations of the same general technique, 
the induction of concept hierarchies.
The similarity between classification trees and clustering trees has also
been noted by Fisher, who points to the 
possibility of using TDIDT (or TDIDT heuristics) in the clustering context
\cite{Fisher93:jrnl} and mentions a few clustering systems that work in a
TDIDT-like fashion \cite{Fisher85:proc}.

Following these views we study top-down induction
of clustering trees. A clustering tree is a decision tree
where the leaves do not contain classes and where each
node as well as each leaf corresponds to a cluster.
To induce clustering trees, we employ 
principles from instance based learning
and decision tree induction.
More specifically, we assume that a distance measure is given that 
computes the distance between two examples. Furthermore, in order 
to compute the distance
between two clusters (i.e. sets of examples), we employ
a function that computes a prototype of a set examples.  
A prototype is then regarded as an example,
which allows to define 
the distance between two clusters 
as the distance between their prototypes.
Given  a distance measure for clusters and the view that each
node of a tree corresponds to a cluster, 
the decision tree algorithm is then adapted to select
in each node the test that will maximize the distance
between the resulting clusters in its subnodes.

Depending on the examples and the distance measure employed one can
distinguish two modes.  In {\em supervised} learning (as in the
classical top-down induction of decision trees paradigm), the distance
measure only takes into account the class information of each example
(see e.g. C4.5 \cite{Quinlan93:other}, CART \cite{Breiman84:other}).
Also, regression trees (SRT \cite{Kramer96:proc}, CART) 
should be considered supervised learning.  In {\em
  unsupervised} learning, the examples may not be classified and the
distance measure does not take into account any class information.
Rather, all attributes or features of the examples are taken into
account in the distance measure.

The Top-down Induction of Clustering trees approach is implemented in
the TIC system.  TIC is a first order clustering system as it does not
employ the classical attribute value representation but that of first
order logical decision trees as in SRT \cite{Kramer96:proc} and \Tilde\
\cite{Blockeel98b:jrnl}. So, the clusters
corresponding to the tree will have first order definitions. On the
other hand, in the current implementation of TIC we only employ
propositional distance measures.

Using TIC we report on a number of experiments. These
experiments demonstrate the power of top-down induction 
of clustering trees. More specifically, we show
that TIC can be used for clustering, for regression,
and for learning classifiers.

This paper significantly expands on an earlier extended 
abstract \cite{DeRaedt97:proc} in that 
TIC now contains  a pruning method
and also that this paper provides new experimental evidence.

This paper is structured as follows.  In Section 2 we discuss the
representation of the data and the induced theories.  Section 3 identifies
possible applications of clustering.  The TIC system
is presented in Section 4.
In Section 5 we empirically evaluate TIC for the proposed applications.
Section 6 presents conclusions and related work.

\section{THE LEARNING PROBLEM}

\subsection{REPRESENTING EXAMPLES}
We employ the {\em learning from interpretations} setting
for inductive logic programming.
For the purposes of this paper, it is sufficient
to regard each example as a small
relational database, i.e. as a set of facts. 
Within learning from interpretations,
one may also specify background knowledge
in the form of a Prolog program which can be used
to derive additional features of the examples.\footnote{The interpretation
  corresponding to each example $e$ is then the minimal Herbrand model
  of $B \wedge e$.}  See \cite{DeRaedt94-AI:jrnl,DeRaedt96-MSL:proc,DeRaedt98:coll}
for more details on learning from interpretations.

For instance, examples for the well-known mutagenesis problem 
\cite{Srinivasan96:jrnl} can be described by interpretations.  Here, an
interpretation is simply an enumeration of 
all the facts we know about one single
molecule: its class, {\em lumo} and {\em logp} values, the atoms and bonds occurring
in it, certain high-level structures\ldots  We can represent 
it e.g.\ as follows:
\{logmutag(-0.7), neg, lumo(-3.025), logp(2.29),
atom(d189\_1,c,22,-0.11), atom(d189\_2,c,22,-0.11),
bond(d189\_1,d189\_2,7),\\
bond(d189\_2,d189\_3,7), \ldots\}


\subsection{FIRST ORDER LOGICAL DECISION TREES}

First order logical decision trees are similar to standard decision trees,
except that the test in each node is a conjunction of literals instead
of an test on an attribute.  They are always binary, as the test can only
succeed or fail.  A detailed discussion of these trees is beyond the scope
of this paper but can be found in \cite{Blockeel98b:jrnl}.
We will use these trees to represent clustering trees.

\begin{figure}
\centering
\epsffile{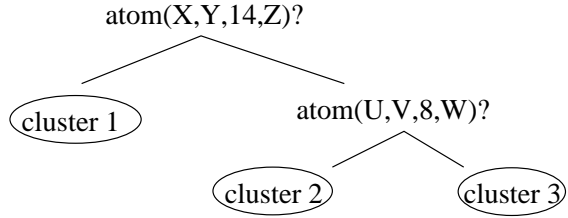}
\caption{A clustering tree}
\label{tree1}
\end{figure}

An example of a clustering tree, in the mutagenesis context,
is shown in Figure \ref{tree1}.  Note that in a classical logical
decision tree leaves would contain classes.  Here, leaves simply
contain sets of examples that belong together.  Also note that variables
occurring in tests are existentially quantified.  The root test, for instance,
tests whether there occurs an atom of type 14 in the molecule.  The whole
set of examples is thus divided into two clusters: a cluster of molecules
containing an atom 14 and a cluster of molecules not containing any.

This view is in correspondence with Langley's viewpoint that a test in
a node is not just a decision criterion, but also a description of the
subclusters formed in this node.  In \cite{Blockeel98b:jrnl} we show
how a logical decision tree can be transformed into an
equivalent logic program, which could alternatively be used to sort
examples into clusters.  The logic program contains invented predicates
that correspond to the clusters.



\subsection{INSTANCE BASED LEARNING AND DISTANCES}

The purpose of conceptual clustering is to obtain clusters such that
intra-cluster distance (i.e.\ the distance between examples belonging
to the same cluster) is as small as possible and the inter-cluster
distance (i.e.\ the distance between examples belonging to different
clusters) is as large as possible.

In this paper, we assume that a distance measure $d$ that computes
the distance $d(e_1,e_2)$ between examples $e_1$ and $e_2$ is given.
Furthermore, there is also a need for measuring the 
distance between different clusters (i.e. between
sets of examples). Therefore we will assume
as well the existence of a prototype function $p$
that computes the prototype $p(E)$ of a  set of examples $E$.
The distance between two
clusters $C_1$ and $C_2$ is then defined as the 
distance $d(p(C_1),p(C_2))$ between the prototypes of the 
clusters. This shows that the prototypes should 
be considered as (possibly) partial
example descriptions. The prototypes should be sufficiently 
detailed as to allow the computation of the distances.

For instance, the distance could be the Euclidean distance $d_1$ between 
the values of one or more numerical attributes,
or it could be the distance  $d_2$ as measured
by a first order distance measure such as used in RIBL \cite{Emde96:proc} or 
KBG \cite{Bisson92:proc} or \cite{Hutchinson97:proc}.

Given the distance at the level of the examples, the principles of
instance based learning can be used to compute the prototypes.  E.g.
$d_1$ would result in a prototype function
$p_1$ that would simply compute the mean for the cluster, whereas
$d_2$ could result in function $p_2$ that would compute the (possibly
reduced) least general generalisation\footnote{Using Plotkin's
[\cite{Plotkin70:coll}] notion of 
$\theta$-subsumption or the variants
corresponding to structural matching
  \cite{Bisson92:proc,Deraedt97-ECML:proc}. } of the examples in the
cluster.

Throughout this paper
we employ only propositional distance measures
and the prototype functions that correspond
to the instance averaging methods 
along the lines of \cite{Langley96:other}.
However, we stress that - in principle - we could 
use any distance measure.
Notice that although we employ only propositional distance measures,
we obtain first order descriptions of the clusters
through the representation of first order logical decision trees. 


\subsection{PROBLEM-SPECIFICATION}

By now we are able to formally specify the clustering problem:\\
\\
{\bf Given}
\begin{itemize}
\item
a set of examples $E$ (each example is a set of tuples in a relational database or equivalently, a set of facts in Prolog),
\item
a background  theory $B$ in the form of a  Prolog program,
\item
a distance measure $d$ that computes the distance
between two examples or prototypes,
\item
a prototype function $p$ that computes the prototype
of a set of examples, 
\end{itemize}
{\bf Find}: a first order clustering tree.\\

Before discussing how this problem can be solved we take a look at possible
applications of clustering trees.
 

\section{APPLICATIONS OF CLUSTERING TREES}

Following Langley's viewpoint, a system such as C4.5
can be considered
a supervised clustering system where the ``distance'' metric is the 
class entropy
within the clusters : lower class entropy within a cluster means that
the examples in that cluster are more similar with respect to their
classes.  Since C4.5 employs class information, it is a supervised
learner.

Clustering can also be done in an unsupervised manner however.  When
making use of a distance metric to form clusters, this distance metric
may or may not use information about the classes of the examples.
Even if it does not use class information, clusters may be coherent
with respect to the class of the examples in them.  

This principle leads to a classification technique that is very
robust with respect to missing class information.  Indeed, even if
only a small percentage of the examples is labelled with a class, one
could perform unsupervised clustering, and assign to each leaf in the
concept hierarchy the majority class in that leaf.  If the
leaves are coherent with respect to classes, this method would yield
relatively high classification accuracy with a minimum of class
information available.  This is quite similar in spirit to Emde's
method for learning from few classified examples, implemented in the
COLA system \cite{Emde94:proc}.

A similar reasoning can be followed for regression, leading to
``unsupervised regression''; again this may be useful in the
case of partially missing information.

We conclude that clustering can extend classification and regression
towards unsupervised learning.  Another extension in the predictive
context is that clusters can be used to predict many or all attributes
of an example at once.

Depending on the application one has in mind, measuring the quality of a
clustering tree is done in different ways.  For classification purposes
predictive accuracy on unseen cases is typically used.  For regression 
an often used criterion is the relative error, which is the mean
squared error of predictions divided by the mean squared error of a
default hypothesis always predicting the mean.
This can be extended towards the clustering context if a distance
measure and prototype function are available: 
\[RE = {{\sum_{i=1}^n d(e_i, \hat{e}_i)^2} \over {\sum_{i=1}^n d(e_i, p)^2}}\]
with $e_i$ the examples, $\hat{e}_i$ the predictions and $p$ the prototype.
(A prediction is, just like a prototype, a partial example description that is
sufficiently detailed to allow the computation of a distance). 
 
If clustering is considered as unsupervised learning of classification or
regression trees, the relative error of only the predicted variable or
the accuracy with which the class variable can be predicted is a suitable
quality criterion.  In this case classes should be available for the
evaluation of the clustering tree, though not during (unsupervised) learning.
Such an evaluation is often done for clusters, see e.g. \cite{Fisher87:jrnl}.

\section{TIC: TOP-DOWN INDUCTION OF CLUSTERING TREES}

A system for top-down induction of clustering trees called TIC has
been implemented as a subsystem of the ILP system \Tilde
\cite{Blockeel98b:jrnl}.  TIC employs the basic
TDIDT framework as it is also incorporated in the \Tilde\ system.  The
main point where TIC and \Tilde\ differ from the propositional TDIDT
algorithm is in the computation of the (first order) tests to be 
placed in a node, see \cite{Blockeel98b:jrnl} for details.
Furthermore, TIC differs from \Tilde\ in that it uses other heuristics
for splitting nodes, an
alternative stopping criterion and alternative tree post-pruning
methods.  We discuss these topics below.

\subsection{SPLITTING}

The splitting criterion used in TIC works as follows.  Given a cluster
$C$ and a test $T$ that will result in two disjoint subclusters $C_1$
and $C_2$ of $C$, TIC computes the distance $d(p(C_1),p(C_2))$, where $p$ is
the prototype function.  The
best test $T$ is then the one that maximizes this distance.  This
reflects the principle that the inter-cluster distance should be as
large as possible.

If the prototype is simply the mean, then maximizing inter-cluster distances 
corresponds to minimizing intra-cluster distances, and
splitting heuristics such as information gain \cite{Quinlan93:other} 
or Gini index \cite{Breiman84:other} can be seen as special cases of
the above principle, as they minimize intra-cluster class diversity.
In the regression context, minimizing intra-cluster variance
(e.g. \cite{Kramer96:proc}) is another instance of this principle.

Note that our distance-based approach has the advantage of being applicable
to both numeric and symbolic data, and thus generalises over regression
and classification.


\subsection{STOPPING CRITERIA}
Stopping criteria are often based on significance tests.  In the
classification context a $\chi^2$-test is often used to check whether
the class distributions in the subtrees differ significantly 
\cite{Clark89:jrnl,Deraedt95-ALT:proc}.
Since regression and clustering use variance as a heuristic for choosing
the best split, a reasonable heuristic for the stopping criterion seems to be
the F-test.  If a set of examples is split into two subsets, the variance
should decrease significantly, i.e.
 
\[F = {{SS/(n-1)} \over {(SS_L+SS_R)/(n-2)}}\]
 
should be significantly large ($SS$ is the sum of squared differences
from the mean inside the set of examples,
 $SS_L$ and $SS_R$ is the same for the two created subsets of the examples,
 $n$ is the total number of examples).\footnote{The F-test is only
theoretically correct for normally distributed populations.  Since this
assumption may not hold, it should here be considered a {\em heuristic} for
deciding when to stop growing a branch, not a real statistical test.}
 
\subsection{PRUNING USING A VALIDATION SET}
The principle of using a validation set to prune trees is very simple.
After using the training set to build a tree, the quality of the
tree is computed on the validation set (predictive accuracy for 
classification trees, inverse of relative error for regression or 
clustering trees).  For each node of the tree the quality of the tree
if it were pruned at that node $Q'$ is compared with the quality
$Q$ of the unpruned tree.  If $Q' > Q$ then the tree is pruned.

Such a strategy has been successfully followed in the context
of classification and regression (e.g. CART \cite{Breiman84:other}) as
well as clustering (e.g. \cite{Fisher96:jrnl}).  Fisher's method
is more complex than ours in that for each individual variable
a different subset of the original tree will be used for prediction.

In the current implementation of \Tilde\ validation set based pruning is
available for all settings.  For clustering and regression it is the only
pruning criterion that is implemented.  It is only reliable for reasonably
large data sets though.  When learning from small data sets performance
decreases because the training set becomes even smaller and with a small
validation set a lot of pruning is due to random influences.


\section{EXPERIMENTS}

\subsection{DATA SETS}

We used the following data sets for our experiments:
\begin{itemize}
\item {\bf Soybeans:} this database \cite{Michalski80:jrnl} contains
  descriptions of diseased soybean plants.  Every plant is described
  by 35 attributes.  A small data set (46 examples, 4 classes) and a
  large one (307 examples, 19 classes) are available at the UCI
  repository \cite{uci:misc}.

\item {\bf Iris:} a simple database of descriptions of iris plants, 
available at the UCI repository.  It contains 3 classes of 50 examples
each. There are 4 numerical attributes.  

\item {\bf Mutagenesis:} this database \cite{Srinivasan96:jrnl}
  contains descriptions of molecules for which the mutagenic activity
  has to be predicted.  Originally mutagenicity was measured by a real
  number, but in most experiments with ILP systems this has been
  discretized into two values (positive and negative).  The database
  is available at the ILP repository \cite{ilprepos:misc}.

  \cite{Srinivasan95:proc}
  introduce four levels of background knowledge; the first 2 contain
  only structural information (atoms and bonds in the molecules), the
  other 2 contain higher level information (attributes describing the
  molecule as a whole and higher level submolecular structures).  For
  our experiments the tests allowed in the trees can make use of
  structural information only (Background 2), though for the
  heuristics numerical information from background 3 can be used.

\item {\bf Biodegradability:} a set of 62 molecules of which
  structural descriptions and molecular weights are given.  The
  biodegradability of the molecules is to be predicted.  This is a
  real number, but has been discretized into four values (fast,
  moderate, slow, resistant) in most past experiments.  The dataset
  was provided to us by S. D{\v z}eroski but is not yet in the public
  domain.

\end{itemize}

The data sets were deliberately chosen to include both propositional
and relational data sets.  For each individual experiment the most
suitable data sets were chosen (w.r.t. size, suitability for a
specific task, and relevant results published in the literature).

Distances were always computed from all numerical attributes,
except when stated otherwise.  For the Soybeans data sets all nominal
attributes were converted into numbers first.



\subsection{EXPERIMENT 1: PRUNING}

In this first experiment we want to evaluate the effect of pruning in
\tich{} on both predictive accuracy and tree complexity.  We have applied
\tich{} to two databases: Soybeans (large version) and Mutagenesis.  We chose
these two because they are relatively large (as noted before, the pruning
strategy is prone to random influences when used with small datasets).

For both data sets tenfold crossvalidations were performed.  In each
run the algorithm divides the learning set in a training set and a
validation set.  Clustering trees are built and pruned in an
unsupervised manner.  The clustering hierarchy before and after
pruning is evaluated by predicting the class of each test example.

\begin{figure}[t]
\begin{minipage}{5cm}
\epsfxsize=6cm
\epsffile{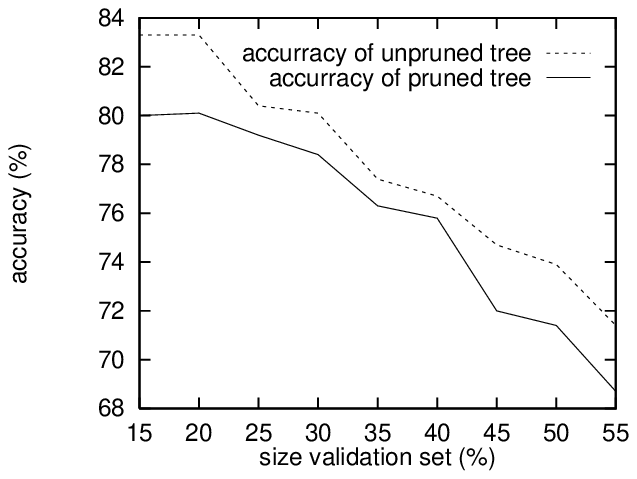}
\end{minipage}
\begin{minipage}{5cm}
\epsfxsize=6cm
\epsffile{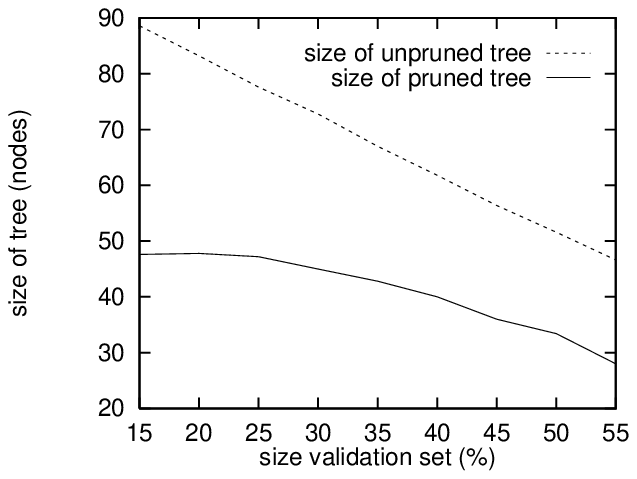}
\end{minipage}
\caption{\label{fig:soyprune_acc}Soybeans: a) Accuracy before and
after pruning; b) number of nodes before and after pruning}
\end{figure}

\begin{figure}[t]
\begin{minipage}{5cm}
\epsfxsize=6cm
\epsffile{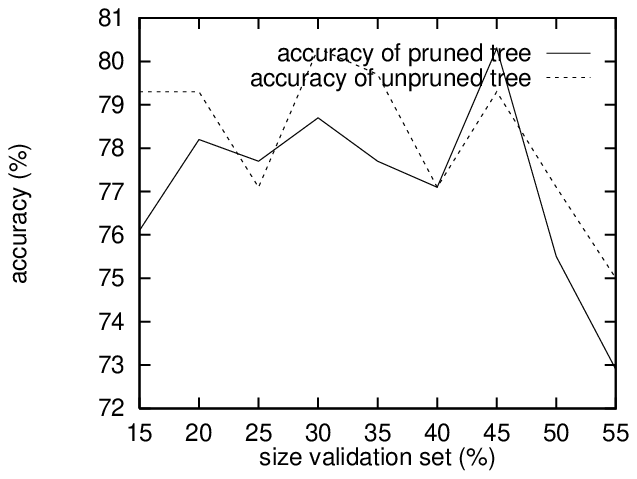}
\end{minipage}
\begin{minipage}{5cm}
\epsfxsize=6cm
\epsffile{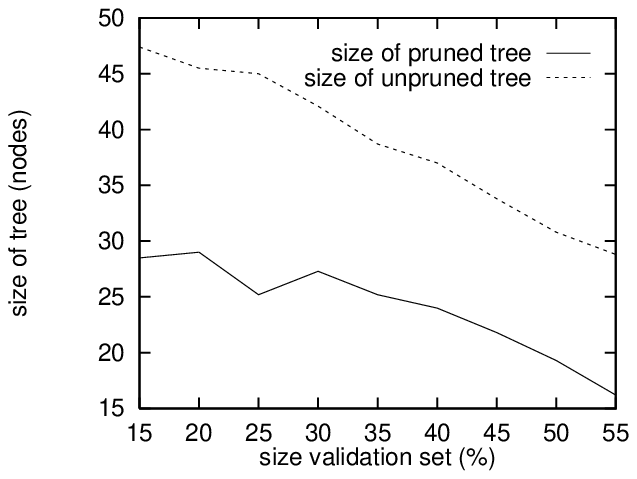}
\end{minipage}
\caption{\label{fig:mutaprune_acc}Mutagenesis: Accuracy and size of the
clustering trees}
\end{figure}

In Figure \ref{fig:soyprune_acc},
the average accuracy of the clustering hierarchies before and
after pruning is plotted against the size of the validation set (this size
is a parameter of TIC), and the same is done for the tree complexity.
The same results for the Mutagenesis database are summarised in 
Figure~\ref{fig:mutaprune_acc}.

From the Soybeans experiment it can be concluded that \Tic's pruning
method results in a slight decrease  in accuracy but a large decrease
in the number of nodes.  The pruning strategy seems relatively stable
w.r.t. the size of the validation set.  The Mutagenesis experiment confirms
these findings (though the decrease in accuracy is less clear here).

\subsection{EXPERIMENT 2: COMPARISON WITH OTHER LEARNERS}

In this experiment we compare \tich{} with propositional clustering
systems and with classification and regression systems.
A comparison with propositional clustering systems is hard to make because
few quantitative results are available in the literature, therefore we also
compare with supervised learners.

We applied \tich{} to the Soybean (small) and Iris databases, performing
tenfold crossvalidations.  Learning is unsupervised, but classes are assumed
to be known at evaluation time (the class of a test example is compared with
the majority class of the leaf the example is sorted into).  Table~\ref{exp2}
compares the results with those obtained with the supervised learner \Tilde.

\begin{table}
\begin{tabular}{|l|l|l|l|l|}
\hline
& \multicolumn{2}{c|}{TIC} & \multicolumn{2}{c|}{\Tilde}\\
\hline
Database & acc. & tree size & acc. & tree size\\
\hline
Soybean & 97\% & 3.9 nodes &  100\% & 3 nodes
\\\hline
Iris   &  92\% & 15 nodes & 94\% & 4 nodes
\\\hline
\end{tabular}
\caption{Comparison of TIC with a supervised learner (averages over 10-fold crossvalidation).}
\label{exp2}
\end{table}

We see that \tich{} obtains high accuracies for these problems.  The only
clustering result we know of is for COBWEB, which obtained 100\%
on the Soybean data set.  This difference is not significant.
\Tilde's accuracies don't differ much from those of \tich{}
which induced the hierarchy without knowledge of the classes.  Tree
sizes are smaller though.

We have also performed an experiment on the Biodegradability 
data set, predicting numbers.  
For this dataset the F-test stopping criterion was used (significance
level 0.01), but no validation set was used given the small size of the
data set.   The distance used is the difference between class values.
Table~\ref{biodeg} compares TIC's performance with \Tilde's
(classification, leave-one-out) and SRT's (regression, sixfold).
\begin{table}
\begin{tabular}{lll}
\hline
l.o.o. \Tilde & classification & acc. = 0.532\\
l.o.o. TIC & regression & RE = 0.740\\
l.o.o. TIC & classif. via regression & acc. = 0.565\\
\hline
6-fold SRT    & regression & RE = 0.34\\
6-fold TIC    & regression & RE = 1.13\\
\hline
\end{tabular}
\caption{Comparison of regression and classification on the biodegradability
data (l.o.o.=leave-one-out).}
\label{biodeg}
\end{table}

Our conclusions are that a) for unsupervised learning TIC performs almost as 
well as other unsupervised or supervised learners, if classification 
accuracy is measured; and b) while there is clearly room for improvement with
respect to using TIC for regression, post-discretization of the regression
predictions shows that this approach is competitive with classical 
approaches to classification.

\subsection{EXPERIMENT 3: PREDICTING MULTIPLE ATTRIBUTES}

Clustering allows to predict multiple attributes.  Since examples in a
leaf must resemble each other as much as possible, attributes must
also agree as much as possible.

By sorting unseen examples down a cluster tree and comparing all attributes
of the example with the prototype attributes, we get an idea of how
good the tree is.  This is an extension of the classical evaluation, as each
attribute in turn is a class now.

We did a tenfold crossvalidation for the following experiment:  using the
training set a clustering tree is induced.  Then, all examples of the
test set are sorted in this hierarchy, and the prediction for all of their
attributes is evaluated.  For each attribute, the value that occurs
most frequently in a leaf is predicted for all test examples sorted in
that leaf.

We used the large soybean database, with pruning.
Table~\ref{soyb2} summarizes the accuracies obtained for each attribute
and compares with the accuracy of majority prediction.
The high accuracies show that most attributes can be predicted very well,
which means the clusters are very coherent.  The mean accuracy of 81.6\%
does not differ significantly from the $83 \pm 2\%$ reported in
\cite{Fisher96:jrnl}.

\begin{table}[t]
\footnotesize
\begin{tabular}{|l|c|c|c|}
\hline {\bf name} & {\bf range} & {\bf default} & {\bf acc.} \\\hline
date            & 0-6 & 21.2\% & 46.3\% \\\hline
plant\_stand    & 0-1 & 52.1\% & 85.0\% \\\hline
precip          & 0-2 & 68.4\% & 79.2\% \\\hline
temp            & 0-2 & 58.3\% & 75.6\% \\\hline
hail            & 0-1 & 68.7\% & 71.3\% \\\hline
crop\_hist      & 0-3 & 32.2\% & 45.0\% \\\hline
area\_damaged   & 0-3 & 32.9\% & 54.4\% \\\hline
severity        & 0-2 & 49.2\% & 63.2\% \\\hline
seed\_tmt       & 0-2 & 45.6\% & 51.1\%\\\hline
germination     & 0-2 & 32.2\% & 45.0\% \\\hline      
plant\_growth   & 0-1 & 65.8\% & 96.4\%\\\hline
leaves          & 0-1 & 89.3\% & 96.4\% \\\hline
leafspots\_halo & 0-2 & 49.5\% & 85.3\%\\\hline
leafspots\_marg & 0-2 & 52.2\% & 86.6\% \\\hline
leafspots\_size & 0-2 & 47.8\% & 87.0\%\\\hline
leaf\_shread    & 0-1 & 75.9\% & 81.4\% \\\hline
leaf\_malf      & 0-1 & 87.3\% & 88.3\%\\\hline
leaf\_mild      & 0-2 & 83.7\% & 88.9\% \\\hline
stem            & 0-1 & 54.1\% & 98.4\%\\\hline
lodging         & 0-1 & 80.7\% & 80.0\% \\\hline     
stem\_cankers   & 0-3 & 58.3\% & 90.6\%\\\hline
canker\_lesion  & 0-3 & 49.1\% & 88.9\% \\\hline
fruiting\_bodies& 0-1 & 73.6\% & 84.3\%\\\hline
external\_decay & 0-2 & 75.6\% & 91.5\% \\\hline
mycelium        & 0-1 & 95.8\% & 96.1\%\\\hline
int\_discolor   & 0-2 & 86.6\% & 95.4\% \\\hline
sclerotia       & 0-1 & 93.2\% & 96.1\%\\\hline
fruit\_pods     & 0-3 & 62.7\% & 91.2\% \\\hline
fruit\_spots    & 0-4 & 53.4\% & 87.0\%\\\hline
seed            & 0-1 & 73.9\% & 85.7\% \\\hline    
mold\_growth    & 0-1 & 80.5\% & 86.6\%\\\hline
seed\_discolor  & 0-1 & 79.5\% & 84.0\% \\\hline
seed\_size      & 0-1 & 81.8\% & 88.6\%\\\hline
shriveling      & 0-1 & 83.4\% & 87.9\% \\\hline
roots           & 0-2 & 84.7\% & 95.8\%\\\hline     
{\bf mean} & & & {\bf 81.6\%}\\\hline
\end{tabular}                                       
\caption{Prediction of all attributes together in the Soybean data set}
\label{soyb2}
\end{table}

\subsection{EXPERIMENT 4: HANDLING MISSING INFORMATION}

It can be expected that clustering, making use of more attributes than just
class attributes, is more robust with respect to missing values.
We showed in Experiment 2 that unsupervised learners (where the
heuristics do not use any class information at all) can yield trees
with predictive accuracies close to those of supervised learners, but
all class information was still available for assigning classes to leaves
after the tree was built.  

In this experiment, we measure the predictive accuracy of trees when class
information as well as other information may be missing, not only for
learning, but also for assigning classes to leaves afterwards, and this for 
several levels of missing information.  Our aim is to investigate how
predictive accuracy deteriorates with missing information, and to compare 
clustering systems that use only class information with systems that use
more information.

We have used the Mutagenesis data set for this experiment (for each example,
there was a fixed probability that the value of a
certain attribute was removed from the data; this probability was increased
for consecutive experiments), comparing the use of only class
information ({\em logmutag}) with the use of three numerical variables
(among which the class) for computing distances.  This experiment is similar
in spirits to the ones performed with COLA \cite{Emde94:proc}.
Table~\ref{muta2} shows the results.
As expected, performance degrades less quickly when more information
is available, which supports the claim that
the use of more than just class information can improve performance in the
presence of missing information.


\begin{table}
\begin{footnotesize}
\begin{tabular}{|c|cc|}
\hline
available numerical data & logmutag &  all three\\
\hline
100\%          & 0.80 &   0.81\\
50\%           & 0.78 &   0.79\\
25\%           & 0.72 &   0.77\\
10\%           & 0.67 &   0.74\\  
\hline
\end{tabular}
\end{footnotesize}
\caption{Classification accuracies obtained for Mutagenesis with several distance functions, and on several levels of missing information.}
\label{muta2}
\end{table}

\section{CONCLUSIONS AND RELATED WORK}

We have presented a novel first order clustering system TIC
within the TDIDT class of algorithms.
TIC integrates ideas from concept-learning
(TDIDT), from instance based learning (the distances
and the prototypes), and from inductive logic programming
(the representations) to obtain a clustering system.
Several experiments were performed that illustrate the type of
tasks TIC is useful for.

As far as related work is concerned,
our work is related to KBG \cite{Bisson92:proc},
which also performs first order clustering.
In contrast to the current version of TIC,
KBG does use a first order similarity measure, which
could also be used within TIC. 
Furthermore, KBG is an agglomerative (bottom-up) clustering algorithm
and TIC a divisive one (top-down). The divisive nature
of TIC makes  TIC  as efficient as classical TDIDT algorithms.   
A final difference with KBG 
is that TIC directly obtains logical descriptions of the 
clusters through the use of the logical decision
tree format. For KBG, these descriptions have to be derived
in a separate step because the clustering process
only produces the clusters (i.e. sets of examples) and not 
their description. 

The instance-based learner RIBL \cite{Emde96:proc} uses an
advanced first order distance metric that might be a good candidate for
incorporation in TIC.

While \cite{Fisher93:jrnl} first made the link between TDIDT and clustering,
our work is inspired mainly by \cite{Langley96:other}.
From this point of view, our work is closely related to
SRT \cite{Kramer96:proc}, who builds regression trees  in a supervised
manner. 
TIC can be considered
a generalization of SRT in that 
TIC can also build trees in an unsupervised manner,
and can predict multiple values.
Finally, we should also refer to a number of 
other approaches to first order clustering, which include 
Kluster \cite{Kietz94-ML:jrnl}, \cite{Yoo91:proc}, \cite{Thompson91:coll}
and \cite{Ketterlin95:proc}.

Future work on TIC includes extending the system so that it can employ
first order distance measures, and investigating the limitations of
this approach (which will require further experiments).

\subsubsection*{Acknowledgements}

Hendrik Blockeel is supported by the Flemish Institute for the
Promotion of Scientific and Technological Research in Industry (IWT).
Luc De Raedt is supported by the Fund for Scientific Research of Flanders.

This work is part of the European Community Esprit
project no. 20237, Inductive Logic Programming 2.
The authors thank Stefan Kramer, who performed the SRT experiments, 
Sa{\v s}o D{\v z}eroski, who provided the Biodegradability database,
Luc Dehaspe and Kurt Driessens for proofreading the paper, and the anonymous
referees for their very valuable comments.



\small

\bibliographystyle{named}
\bibliography{.mlbib}

\end{document}